\documentclass[11pt,a4paper]{article}
\usepackage[hyperref]{emnlp-ijcnlp-2019}
\usepackage{times}
\usepackage{latexsym}
\usepackage{graphicx}
\usepackage{amsmath,bm}
\usepackage{mathrsfs}
\usepackage{multirow}
\usepackage{CJKutf8}
\usepackage{verbatim}
\usepackage{amssymb}
\usepackage{algorithm}
\usepackage{algorithmic}
\usepackage{tabu}
\usepackage{booktabs}
\usepackage{color}
\usepackage{subfigure}

\usepackage{url}

\aclfinalcopy 

\title{ARAML: A Stable Adversarial Training Framework for Text Generation}

\author{Pei Ke\thanks{\quad Equal contribution}, Fei Huang$^*$, Minlie Huang\thanks{\quad Corresponding author: Minlie Huang}, Xiaoyan Zhu \\
Institute for Artificial Intelligence, 
State Key Lab of Intelligent Technology and Systems \\
Beijing National Research Center for Information Science and Technology \\
Department of Computer Science and Technology, Tsinghua University, Beijing 100084, China \\
  {\tt kepei1106@outlook.com, f-huang18@mails.tsinghua.edu.cn} \\
  {\tt aihuang@tsinghua.edu.cn, zxy-dcs@tsinghua.edu.cn} \\
  }

\date{}

\begin{document}
\maketitle

\begin{CJK*}{UTF8}{gbsn}

\begin{abstract}
Most of the existing generative adversarial networks (GAN) for text generation suffer from the instability of reinforcement learning training algorithms such as policy gradient, leading to unstable performance.
To tackle this problem, we propose a novel framework called Adversarial Reward Augmented Maximum Likelihood (ARAML).
During adversarial training, the discriminator 
assigns rewards to samples which are acquired from a stationary distribution near the data rather than the generator's distribution.
The generator is optimized with maximum likelihood estimation augmented by the discriminator's rewards instead of policy gradient. 
Experiments show that our model can outperform state-of-the-art text GANs with a more stable training process.
\end{abstract}

\section{Introduction}

Natural text generation, as a key task in NLP, has been advanced substantially thanks to the flourish of neural models \cite{bengiolanguagemodel2003, mikolov2010rnn}. Typical frameworks such as sequence-to-sequence (seq2seq) have been applied to various generation tasks, including machine translation \cite{sutskever2014sequence} and dialogue generation \cite{Vinyals2015A}. The standard paradigm to train such neural models is maximum likelihood estimation (MLE), which maximizes the log-likelihood of observing each word in the text given the ground-truth proceeding context \cite{graves2013rnn}.

Although widely used, MLE suffers from the exposure bias problem \cite{bengio2015schedule,sentencelevel2016}: during test, the model sequentially predicts the next word conditioned on its previous generated words while during training conditioned on ground-truth words. To tackle this problem, generative adversarial networks (GAN) with reinforcement learning (RL) training approaches have been introduced to text generation tasks~\cite{yu2017seqgan,che2017maligan,lin2017rankgan,fedus2018maskgan,guo2018leakgan,shi2018irl,xu2018dpgan}, where the discriminator is trained to distinguish real and generated text samples to provide reward signals for the generator, and the generator is optimized via policy gradient \cite{yu2017seqgan}.

However, recent studies have shown that potential issues of training GANs on discrete data are more severe than exposure bias \cite{ganevaluation2018, ganfallingshort2018}.
One of the fundamental issues when generating discrete text samples with GANs is training instability.
Updating the generator with policy gradient always leads to an unstable training process because it's difficult for the generator to derive positive and stable reward signals from the discriminator even with careful pre-training \cite{che2017maligan}.
As a result, the generator gets lost due to the high variance of reward signals and the training process may finally collapse \cite{li2017adversarial}.

In this paper, we propose a novel adversarial training framework called Adversarial Reward Augmented Maximum Likelihood (ARAML) to deal with the instability issue of training GANs for text generation.
At each iteration of adversarial training, we first train the discriminator to assign higher rewards to real data than to generated samples. Then, inspired by reward augmented maximum likelihood (RAML) \cite{Norouzi2016raml}, the generator is updated on the samples acquired from a stationary distribution with maximum likelihood estimation (MLE), weighted by the discriminator's rewards. This stationary distribution is designed to guarantee that training samples are surrounding the real data, thus the exploration space of our generator is indeed restricted by the MLE training objective, resulting in more stable training.
Compared to other text GANs with RL training techniques, our framework acquires samples from the stationary distribution rather than the generator's distribution, and uses RAML training paradigm to optimize the generator instead of policy gradient. Our contributions are mainly as follows:
\begin{itemize}
    \item We analyze the fundamental issue of current GANs for text generation from the perspectives of training instability.
    \item We propose a novel framework called Adversarial Reward Augmented Maximum Likelihood (ARAML), which incorporates stable RAML training into adversarial training paradigm. Experimental results on three text generation tasks show the effectiveness of our method.
\end{itemize}

\section{Related Work}

Recently, text generation has been widely studied with neural models trained with maximum likelihood estimation \cite{graves2013rnn}. However, MLE tends to generate universal text \cite{li2016diversityMMI}. Various methods have been proposed to 
enhance the generation quality by refining the objective function \cite{li2016diversityMMI,mou2016sequenceBackwardForward} or modifying the generation distribution with external information like topic \cite{xing2017topicAware}, sentence type \cite{ke2018senfunc}, emotion \cite{ecm2018} and knowledge \cite{zhou2018commonsense}.

As mentioned above, MLE suffers from the exposure bias problem \cite{bengio2015schedule,sentencelevel2016}. Thus, reinforcement learning has been introduced to text generation tasks such as policy gradient \cite{sentencelevel2016} and actor-critic \cite{actorcritic2017}. \cite{Norouzi2016raml} proposed an efficient and stable approach called Reward Augmented Maximum Likelihood (RAML), which connects the log-likelihood and expected rewards to incorporate MLE training objective into RL framework.

Since some text generation tasks have no explicit metrics to be directly optimized, adversarial training has been applied to generating discrete text samples with a discriminator to learn a proper reward. For instance, SeqGAN \cite{yu2017seqgan} devised a discriminator to distinguish the real data and generated samples, and a generator to maximize the reward from the discriminator via policy gradient. Other variants of GANs have been proposed to improve the generator or the discriminator. To improve the generator, MaliGAN \cite{che2017maligan} developed a normalized maximum likelihood optimization target for the generator to stably model the discrete sequences. LeakGAN \cite{guo2018leakgan} guided the generator with reward signals leaked from the discriminator at all generation steps to deal with long text generation task. MaskGAN \cite{fedus2018maskgan} employed an actor-critic architecture to make the generator fill in missing text conditioned on the surrounding context, which is expected to mitigate the problem of mode collapse.
As for the discriminator, RankGAN \cite{lin2017rankgan} replaced traditional discriminator with a ranker to learn the relative ranking information between the real texts and generated ones. Inverse reinforcement learning \cite{shi2018irl} used a trainable reward approximator as the discriminator to provide dense reward signals at each generation step. DPGAN \cite{xu2018dpgan} introduced a language model based discriminator and regarded cross-entropy as rewards to promote the diversity of generation results.

The most similar works to our model are RAML \cite{Norouzi2016raml} and MaliGAN \cite{che2017maligan}: 1) Compared with RAML, our model adds a discriminator to learn the reward signals instead of choosing existing metrics as rewards. We believe that our model can adapt to various text generation tasks, particularly those without explicit evaluation metrics. 2) Unlike MaliGAN, we acquire samples from a fixed distribution near the real data rather than the generator's distribution, which is expected to make the training process more stable.

\begin{figure*}[!htp]
  \centering
  \includegraphics[width=1.0\linewidth]{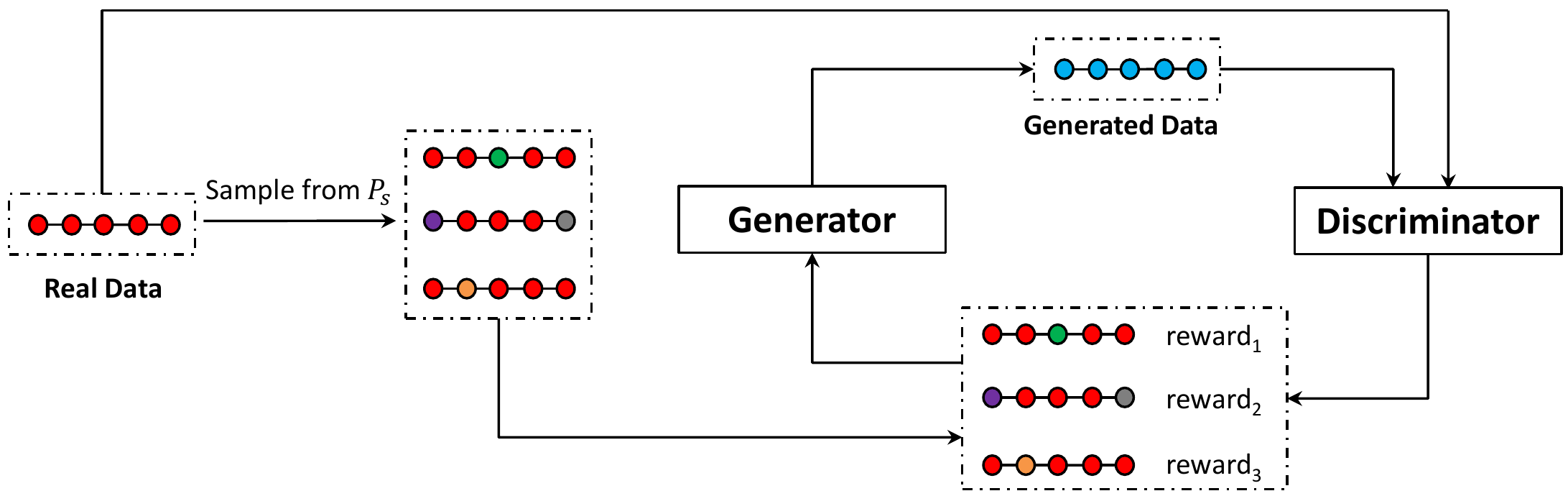}
  \caption{Overview of ARAML. The training samples are acquired from a stationary distribution $P_s$ based on the real data. The generator is then trained on the samples augmented by the discriminator's rewards. The discriminator is trained to distinguish real data and generated data. }
  \label{fig:overview}
\end{figure*}

\section{Model}

\subsection{Task Definition and Model Overview}

Text generation can be formulated as follows: given the real data distribution $P_{\mathrm{data}}(X)$, the task is to train a generative model $G_\theta$ where $P_{G_\theta}(X)$ can fit $P_{\mathrm{data}}(X)$ well. In this formulation, $X=x_1 x_2 \cdots x_m$ and $x_t (1\leq t\leq m)$ denotes a word in the vocabulary $\mathcal{V}$.

Figure \ref{fig:overview} shows the overview of our model ARAML. This adversarial training framework consists of two phases:
1) The discriminator is trained to assign higher rewards to real data than to generated data.
2) The generator is trained on the samples acquired from a stationary distribution with reward augmented MLE training objective.
This training paradigm of the generator indeed constrains the search space with the MLE training objective, which alleviates the issue of unstable training.

\subsection{Discriminator}

The discriminator $D_{\phi}$ aims to distinguish real data and generated data like other GANs. Inspired by Least-Square GAN \cite{mao2017lsgan}, we devise the loss function as follows:
\begin{multline}
\mathcal{L}_{D_\phi} = \frac{1}{2}\mathbb{E}_{X\sim P_{\mathrm{data}}(X)}\left[(D_{\phi}(X)-1)^2\right] \\
+\frac{1}{2}\mathbb{E}_{X\sim P_{G_\theta}(X)}\left[(D_{\phi}(X))^2\right] \label{eqn:dis}
\end{multline}
This loss function forces the discriminator to assign higher rewards to real data than to generated data, so the discriminator can learn to provide more proper rewards as the training proceeds.

\subsection{Generator}

The training objective of our generator $G_\theta$ is derived from the objective of other discrete GANs with RL training method:
%
\begin{align}
    \mathcal{L}_{\mathrm{RL},\theta} & = - \mathbb{E}_{X \sim P_{G_{\theta}}(X)}[r_\phi(X)] - \tau \mathbb{H} (P_{G_\theta}(X))
\end{align}  
where $r_\phi(X)$ denotes the rewards from the discriminator $D_{\phi}$ and the entropy regularized term $\mathbb{H} (P_{G_\theta}(X))$ encourages $G_{\theta}$ to generate diverse text samples. $\tau$ is a temperature hyper-parameter to balance these two terms.

As mentioned above, discrete GANs suffer from the instability issue due to policy gradient, thus they are consequently difficult to train.
Inspired by RAML \cite{Norouzi2016raml}, we introduce an exponential payoff distribution $Q_\phi(X)$ to connect RL loss with RAML loss:
\begin{align}
Q_\phi(X) = \frac{1}{Z} \exp(r_\phi(X)/\tau)
\end{align}
where $Z = \sum_X{\exp(r_\phi(X)/\tau)}$. Thus, we can rewrite $\mathcal{L}_{\mathrm{RL},\theta}$ with $P_{G_\theta}(X)$ and $Q_\phi(X)$ as follows:
\begin{align}
    \mathcal{L}_{\mathrm{RL},\theta} = {\tau}KL(P_{G_{\theta}}(X) || Q_\phi(X)) + constant
\end{align}  
Following RAML, we remove the constant term and optimize the KL divergence in the opposite direction:
\begin{align}
    \mathcal{L}&_{\mathrm{RAML},\theta} = KL(Q_\phi(X) || P_{G_{\theta}}(X)) \notag \\
    & = - \mathbb{E}_{X \sim Q_\phi(X)}[\log P_{G_{\theta}}(X)] - \mathbb{H}(Q_\phi(X)) \notag \\
    & = - \mathbb{E}_{X \sim Q_\phi(X)}[\log P_{G_{\theta}}(X)] + constant \label{eqn:raml}
\end{align}
where $\mathbb{H}(Q_\phi(X))$ is a constant in the training phase of the generator. It has been proved that $\mathcal{L}_{\mathrm{RL},\theta}$ and $\mathcal{L}_{\mathrm{RAML},\theta}$ are equivalent up to their first order Taylor approximations, and they have the same global optimum \cite{Norouzi2016raml}.
$\mathcal{L}_{\mathrm{RAML},\theta}$ can be trained in a MLE-like fashion but sampling from the distribution $Q_\phi(X)$ is intractable in the adversarial setting, because $Q_\phi(X)$ varies with the discriminator $D_\phi$.
Thus, we introduce importance sampling to separate sampling process from $D_\phi$ and obtain the final loss function:
\begin{align}
    \mathcal{L}_{G_\theta} = - \mathbb{E}_{X\sim P_s(X)}[W_\phi(X) \log P_{G_{\theta}}(X)] \label{eqn:generator}
\end{align}
where $P_s(X)$ denotes a stationary distribution and $W_\phi(X) \propto Q_\phi(X) / P_s(X) $. To optimize this loss function, we first construct the fixed distribution $P_s(X)$ to get samples, and 
devise the proper reward function $r_\phi(X)$ to train the generator in a stable and effective way.

\subsubsection{Sampling}
\label{sec:editdis}
We construct the distribution $P_s$ based on $P_{\mathrm{data}}$:
\begin{align}
    P_s(X) = \mathbb{E}_{X\sim P_{\mathrm{data}}(X)}[ P_s(X_s|X)] \label{eqn:P_s}
\end{align}
In this way, $P_s(X_s|X)$ can be designed to guarantee that $P_s(X)$ is near $P_{\mathrm{data}}(X)$, leading to a more stable training process. To obtain a new sample $X_s$ from a real data sample $X$, we can design three steps which contain sampling an edit distance $d$, the positions $\{p_1,p_2,\cdots,p_d\}$ for substitution and the new words $\{w_1,w_2,\cdots,w_d\}$ filled into the corresponding positions.
Thus, $P_s(X_s|X)$ can be decomposed into three terms:
\begin{multline}
    P_s(X_s|X) = P(d,p,w|X) \\
    = P(d|X)P(p|X,d)P(w|X,d,p)
\end{multline}

The first step is to sample an edit distance based on a real data sample $X$, where $X=x_1x_2\cdots x_m$ is a sequence of length $m$. The number of sentences which have the edit distance $e$ to some input sentence can be computed approximately as below:
\begin{equation}
    c(e,m)=\binom{m}{e}\cdot\left(|\mathcal{V}|-1\right)^e
\end{equation}
where $c(e,m)$ denotes the number of sentences which have an edit distance $e(e\in \{0,1,...,m\})$ to a sentence of length $m$, and $|\mathcal{V}|$ indicates the size of vocabulary. 
We then follow \cite{Norouzi2016raml} to re-scale the counts by $\exp\{-e/\tau\}$ and do normalization, so that we can sample an edit distance $d^*$ from:
\begin{equation}
    P(d=d^*|X)
    = \frac{\exp\{-d^*/\tau\}c(d^*,m)}{\sum_{e=0}^m \exp\{-e/\tau\}c(e,m)} \label{eqn:dist}
\end{equation}
where $\tau$, as a temperature hyper-parameter, restricts the search space surrounding the original sentence. Larger $\tau$ brings more samples with long edit distances.

The next step is to select positions for substitution based on the sampled edit distance $d^*$. Intuitively, we can randomly choose $d^*$ distinct positions in $X$ to be replaced by new words. The probability of choosing the position $p^*$ is calculated as follows:
\begin{equation}
    P(p=p^*|X,d=d^*) = \frac{d^*}{m} \label{eqn:possampling}
\end{equation}
Following this sampling strategy, we can obtain the position set $\{p_1,p_2,\cdots,p_{d^*}\}$. This strategy approximately guarantees that the edit distance between a new sentence and the original sentence is $d^*$.

At the final step, our model determines new words for substitution at each sampled position $p_j(j=1,2,...,d^*)$. We can formulate this sampling process from the original sequence $X$ to a new sample $X_s$ as a sequential transition $X=X_0\to X_1\to\cdots\to X_{d^*}=X_s$. At each step from $X_{j-1}$ to $X_j$ $(j=1,\cdots,d^*)$, we first sample a new word $w_j$ from the distribution $P(w|X_{j-1},p=p_j)$, then replace the old word at position $p_j$ of $X_{j-1}$ to obtain $X_j$. The whole sampling process can be decomposed as follows:
\begin{align}
    P(w|X,d= d^*,& p=\{p_1,p_2,\cdots,p_{d^*}\}) \notag \\ 
    =&\prod_{j=1}^{d^*} P(w_j|X_{j-1},p=p_j) \label{eqn:wordsample}
\end{align}
There are two common sampling strategies to model $P(w|X_{j-1},p=p_j)$, i.e. random sampling and constrained sampling. Random sampling strategy samples a new word $w_j$ according to the uniform distribution over the vocabulary $\mathcal{V}$ \cite{Norouzi2016raml}, while constrained sampling strategy samples $w_j$ to maximize the language model score of the target sentence $X_j$ \cite{su2018gibbssampling, miao2019cgmh}. Here, we adopt constrained sampling in our model and compare the performances of two strategies in the experiment.

\subsubsection{Training}

We devise the reward function $r_\phi(X)$ according to the discriminator's output $D_\phi(X)$ and the stationary distribution $P_s(X)$:
\begin{align}
    r_\phi(X) = \tau\cdot \left[\log P_s(X) + D_\phi(X)\right]
\end{align}
Intuitively, this reward function encourages the generator to generate sentences with large sampling probability and high rewards from the discriminator.
%
Thus, the weight of samples $W_\phi(X)$ can be calculated as follows:
\begin{align}
    W_\phi(X) \propto \frac{Q_\phi(X)}{P_s(X)} \propto \exp\left\{D_\phi(X)\right\}
\end{align}
So far, we can successfully optimize the generator's loss $\mathcal{L}_{G_\theta}$ via Equation \ref{eqn:generator}.
This training paradigm makes our generator avoid possible variances caused by policy gradient and get more stable reward signals from the discriminator, because our generator is restricted to explore the training samples near the real data.

\begin{algorithm}[htb] 
\caption{Adversarial Reward Augmented Maximum Likelihood} 
\label{alg:araml} 
\begin{algorithmic}[1] 
\REQUIRE ~~\\ 
Total adversarial training iterations: $N\_iters$\\
Steps of training generator: $G\_steps$\\
Steps of training discriminator: $D\_steps$ \\
\STATE Pre-train the generator $G_\theta$ with MLE loss
\STATE Generate samples from $P_{G_\theta}$
\STATE Pre-train the discriminator $D_\phi$ via Eq.(\ref{eqn:dis})
\STATE Construct $P_s(X)$ via Eq.(\ref{eqn:P_s}) - Eq.(\ref{eqn:wordsample})
\FOR{each $s = 1,2,...,N\_iters$}
\FOR{each $j = 1,2,...,G\_steps$}
\STATE Update $G_\theta$ via Eq.(\ref{eqn:generator})
\ENDFOR
\FOR{each $k = 1,2,...,D\_steps$}
\STATE Update $D_\phi$ via Eq.(\ref{eqn:dis})
\ENDFOR
\ENDFOR
\end{algorithmic}
\end{algorithm}

\subsection{Extension to Conditional Text Generation}

We have shown our adversarial training framework for text generation tasks without an input. Actually, it can also be extended to conditional text generation tasks like dialogue generation. Given the data distribution $P_{\mathrm{data}}(C,X)$ where $C,X$ denote contexts and responses respectively, the objective function of ARAML's generator can be modified as below:
\begin{align}
\mathcal{L}&_{G_{\theta}} = - \mathbb{E}_{\left(C, X\right) \sim P_{\mathrm{data}}(C,X)} \big[ \notag \\
& \mathbb{E}_{X_s \sim P_s(X_s | C,X)} \left[ W_{\phi}(C,X_s) \log P_{G_{\theta}}(X_s | C) \right] \big]
\label{eqn:ARAML_dialog}
\end{align}
where $W_{\phi}(C,X_s) \propto \exp\{D_{\phi}(C,X_s)\}$ and $D_{\phi}(C,X_s)$ is trained to distinguish whether $X_s$ is the true response to $C$.


\subsection{Comparison with RAML and MaliGAN}
The most similar works to our framework are RAML \cite{Norouzi2016raml} and MaliGAN \cite{che2017maligan}. The main difference among them is the training objective of their generators. We have shown different objective functions in Table \ref{tab:comparision}. For comparison, we use the form with no input for all the three models.

Our model is greatly inspired by RAML, which gets samples from a non-parametric distribution $Q(X)$ constructed based on a specific reward. Compared to RAML, our reward comes from a learnable discriminator which varies as the adversarial training proceeds rather than a specific reward function. This difference equips our framework with the ability to adapt to the text generation tasks with no explicit evaluation metrics as rewards.

Our model is also similar to MaliGAN, which gets samples from the generator's distribution. In MaliGAN's training objective, $G_{\theta'}$ also indicates the generator's distribution but it's used in the sampling phase and fixed at each optimization step. The weight of samples $W^{'}_\phi(X) \propto \frac{D_{\phi}(X)}{1-D_{\phi}(X)}$. Different from our model, MaliGAN acquires samples from the generator's distribution $P_{G_{\theta'}}$, which usually brings samples with low rewards even with careful pre-training for the generator, leading to training instability. Instead, our framework gets samples from a stationary distribution $P_s$ around real data, thus our training process is more stable.

\begin{table} [!htp]
\centering
\small
\setlength{\tabcolsep}{1.0mm}{
\begin{tabular}{cc}
\hline
Model  & Training Objective of Generator \\
\hline
RAML  & $\mathcal{L}_{G_\theta} = - \mathbb{E}_{X \sim Q(X)} [\log P_{G_{\theta}}(X)]$ \\
MaliGAN & $\mathcal{L}_{G_\theta} = - \mathbb{E}_{X\sim P_{G_{\theta'}}(X)}[W^{'}_\phi(X) \log P_{G_{\theta}}(X)]$ \\ 
ARAML  & $\mathcal{L}_{G_\theta} = - \mathbb{E}_{X\sim P_s(X)}[W_\phi(X) \log P_{G_{\theta}}(X)]$ \\
\hline
\end{tabular}}
\caption{Training objectives of generators for RAML, MaliGAN and ARAML.}
\label{tab:comparision}
\end{table}

\section{Experiment}

\subsection{Datasets}
\begin{table} [!htp]
\centering
\small
\setlength{\tabcolsep}{1.0mm}{
\begin{tabular}{cccc}
\hline
Dataset  & Amount(Train/Test) &  Vocabulary & Length \\
\hline
COCO & 80,000/5,000 & 4,839 & 12.8 \\
EMNLP2017 & 49,996/10,000 & 5,721 & 27.8 \\
WeiboDial & 100,000/5,000 & 7,998 & 7.3/10.8 \\ 
\hline
\end{tabular}}
\caption{Statistics of COCO, EMNLP2017 WMT and WeiboDial. The average lengths 7.3/10.8 of WeiboDial indicate the lengths of posts and responses, respectively.}
\label{tab:datastat}
\end{table}

We evaluated ARAML on three datasets: COCO image caption dataset \cite{chen2015coco}, EMNLP2017 WMT dataset\footnote{http://statmt.org/wmt17/translation-task.html} and WeiboDial single-turn dialogue dataset \cite{qian2018personality}. COCO and EMNLP2017 WMT are the common benchmarks with no input to evaluate the performance of discrete GANs, and we followed the existing works to preprocess these datasets \cite{shi2018irl,guo2018leakgan}. WeiboDial, as a dialogue dataset, was applied to test the performance of our model with input trigger. We simply removed post-response pairs containing low-frequency words and randomly selected a subset for our training/test set.
The statistics of three datasets are presented in Table \ref{tab:datastat}.


\subsection{Baselines}

We compared our model with MLE, RL and GAN baselines. Since
COCO and EMNLP2017 WMT don't have input while WeiboDial regards posts as input, we chose the following baselines respectively: 

\noindent \textbf{MLE}: a RNN model trained with MLE objective \cite{graves2013rnn}. Its extension, Seq2Seq, can work on the dialogue dataset \cite{sutskever2014sequence}.

\noindent \textbf{SeqGAN}: The first text GAN model that updates the generator with policy gradient based on the rewards from the discriminator \cite{yu2017seqgan}.

\noindent \textbf{LeakGAN}: A variant of SeqGAN that provides rewards based on the leaked information of the discriminator for the generator
\cite{guo2018leakgan}.

\noindent \textbf{MaliGAN}: A variant of SeqGAN that optimizes the generator with a normalized maximum likelihood objective \cite{che2017maligan}.

\noindent \textbf{IRL}: This inverse reinforcement learning method replaces the discriminator with a reward approximator to provide dense rewards 
\cite{shi2018irl}.

\noindent \textbf{RAML}: A RL approach to incorporate MLE objective into RL training framework, which regards BLEU as rewards \cite{Norouzi2016raml}.

\noindent \textbf{DialogGAN}: An extension of SeqGAN tuned to dialogue generation task with MLE objective added to the adversarial objective \cite{li2017adversarial}. 

\noindent \textbf{DPGAN}: A variant of DialogGAN which uses a language model based discriminator and regards cross-entropy as rewards \cite{xu2018dpgan}.

Note that MLE, SeqGAN, LeakGAN, MaliGAN and IRL are the baselines on COCO and EMNLP2017 WMT, while MLE, RAML, DialogGAN, and DPGAN on WeiboDial. The original codes are used to test the baselines.

\begin{table} [!htp]
\centering
\small
\setlength{\tabcolsep}{0.6mm}{
\begin{tabular}{l|c|c}
\hline
Dataset & COCO / EMNLP2017 &  WeiboDial \\
\hline
Generator & LSTM  & GRU \\
\hline
Discriminator & GRU \& CNN & GRU \& MLP \\
\hline
\multirow{2}*{Temperature} & 0.85 (COCO) & \multirow{2}*{0.95} \\
& 0.9 (EMNLP2017) & \\
\hline
Sampling size & 5 & 5 \\
\hline
Dimension of & \multirow{2}*{128} & \multirow{2}*{100} \\
word embedding & & \\
\hline
Batch size & 100 & 100 \\
\hline
Pretraining epochs & \multirow{2}*{50 / 15 / 50} & \multirow{2}*{50 / 10 / 30} \\
(G/D/LM) & & \\
\hline
Optimizer & Adam & Adam \\
\hline
Learning rate(G/D) & 0.001 / 0.0001 & 0.001 / 0.0001 \\
\hline
\end{tabular}}
\caption{Implementation details of ARAML. G/D/LM indicates the generator / discriminator / language model used in constrained sampling, respectively.}
\label{tab:implementation}
\end{table}

\subsection{Implementation Details}

The implementation details of our model are shown in Table \ref{tab:implementation}. For COCO / EMNLP2017, the generator is a LSTM unit \cite{Hochreiter1997LSTM} with 128 cells, and the discriminator is implemented based on \cite{yu2017seqgan}. For WeiboDial, the generator is an encoder-decoder structure with attention mechanism, where both the encoder and the decoder consist of a two-layer GRU \cite{Cho2014GRU} with 128 cells. The discriminator is implemented based on \cite{tao2018ruber}. The language model used in the constrained sampling of ARAML is implemented in the same setting as the generators, and is pre-trained on the training set of each dataset. The codes and the datasets are available at \url{https://github.com/kepei1106/ARAML}.

As for the details of the baselines, the generators of all the baselines except LeakGAN are the same as ours. Note that the generator of LeakGAN consists of a hierarchical LSTM unit, thus we followed the implementation in the original paper. In terms of the differences, the discriminators of GAN baselines are implemented based on the original papers. Other hyper-parameters of baselines including batch size, learning rate, and pre-training epochs, were set based on the original codes, because the convergence of baselines is sensitive to these hyper-parameters.

\begin{table*} [!htp]
\centering
\tiny
\setlength{\tabcolsep}{1.0mm}{
\begin{tabular}{c|ccc|ccc}
\hline
\multirow{2}*{Model} & \multicolumn{3}{c|}{COCO}  & \multicolumn{3}{c}{EMNLP2017 WMT} \\
\cline{2-7}
& PPL-F & PPL-R  & S-BLEU-2/3/4 & PPL-F & PPL-R  & S-BLEU-2/3/4 \\
\hline
MLE &  18.83 $\pm$ 0.51 & 38.81 $\pm$ 0.97 &0.790 $\pm$ 0.006 / 0.598 $\pm$ 0.009 / 0.419 $\pm$ 0.010 & 55.64 $\pm$ 1.03 & 192.33 $\pm$ 9.51 &0.771 $\pm$ 0.005 / 0.505 $\pm$ 0.009 / 0.304 $\pm$ 0.008  \\
SeqGAN & 33.07 $\pm$ 1.98 & 49.24 $\pm$ 2.36 & 0.814 $\pm$ 0.005 / 0.619 $\pm$ 0.008 / 0.430 $\pm$ 0.007 & 67.60 $\pm$ 3.48 & 276.82 $\pm$ 5.12 & 0.786 $\pm$ 0.019 / 0.500 $\pm$ 0.029 / 0.276 $\pm$ 0.023 \\
LeakGAN & \textbf{11.43 $\pm$ 2.74} & 87.54 $\pm$ 6.42 & 0.877 $\pm$ 0.032 / 0.762 $\pm$ 0.045 / 0.645 $\pm$ 0.049 &  \textbf{17.92 $\pm$ 1.77} & 491.70 $\pm$ 18.29 & 0.890 $\pm$ 0.013 / 0.751 $\pm$ 0.011 / 0.604 $\pm$ 0.009 \\
MaliGAN  & 47.16 $\pm$ 2.94 & 54.40 $\pm$ 1.29 & 0.786 $\pm$ 0.005 / 0.572 $\pm$ 0.008 / 0.370 $\pm$ 0.007 & 126.84 $\pm$ 11.18 & 288.20 $\pm$ 16.48 & 0.780 $\pm$ 0.019 / 0.494 $\pm$ 0.032 / 0.265 $\pm$ 0.028 \\
IRL & 41.86 $\pm$ 11.82 & 84.23 $\pm$ 7.02 & 0.857 $\pm$ 0.014 / 0.687 $\pm$ 0.031 / 0.499 $\pm$ 0.062 &  285.20 $\pm$ 36.47 & 1124.57 $\pm$ 109.80 & 0.890 $\pm$ 0.008 / 0.656 $\pm$ 0.052 / 0.406 $\pm$ 0.077 \\
\hline
ARAML & 26.97 $\pm$ 0.55 & \textbf{35.79 $\pm$ 0.49} & \textbf{0.777 $\pm$ 0.005} / \textbf{0.560 $\pm$ 0.006}/ \textbf{0.366 $\pm$ 0.008} & 77.90 $\pm$ 0.70 & \textbf{188.41 $\pm$ 3.18} & \textbf{0.745 $\pm$ 0.002} / \textbf{0.455 $\pm$ 0.006} / \textbf{0.257 $\pm$ 0.006} \\
\hline
\end{tabular}}
\caption{Automatic evaluation on COCO and EMNLP2017 WMT. Each metric is presented with mean and standard deviation.}
\label{tab:cocoauto}
\end{table*}

\subsection{Language Generation on COCO and EMNLP2017 WMT}
We adopted forward/reverse perplexity \cite{zhao2018arae} and Self-BLEU \cite{zhu2018texygen} to evaluate the quality of generated texts.
Forward perplexity (PPL-F) indicates the perplexity on the generated data provided by a language model trained on real data to measure the fluency of generated samples. Reverse perplexity (PPL-R) switches the roles of generated data and real data to reflect the discrepancy between the generated distribution and the data distribution.
Self-BLEU (S-BLEU) regards each sentence in the generated collection as hypothesis and the others as reference to obtain BLEU scores, which evaluates the diversity of generated results.

Results are shown in Table \ref{tab:cocoauto}. LeakGAN performs best on forward perplexity because it can generate more fluent samples. 
As for reverse perplexity, our model ARAML beats other baselines, showing that our model can fit the data distribution better. Other GANs, particularly LeakGAN, obtain high reverse perplexity due to mode collapse \cite{shi2018irl}, thus they only capture limited fluent expressions, resulting in large discrepancy between the generated distribution and data distribution.
ARAML also outperforms the baselines in terms of Self-BLEU, indicating that our model doesn't fall into mode collapse with the help of the MLE training objective and has the ability to generate more diverse sentences.

We also provide standard deviation of each metric in Table \ref{tab:cocoauto}, reflecting the stability of each model's performance. Our model ARAML nearly achieves the smallest standard deviation in all the metrics, indicating that our framework outperforms policy gradient in the stability of adversarial training.

\begin{table*} [!htp]
\centering
\small
\setlength{\tabcolsep}{1.0mm}{
\begin{tabular}{l|ccc|c|ccc|c}
\hline
\multirow{2}*{Model} & \multicolumn{3}{c|}{Grammaticality} & \multirow{2}*{$\kappa$} & \multicolumn{3}{c|}{Relevance} & \multirow{2}*{$\kappa$} \\
\cline{2-4} \cline{6-8}
& Win (\%) & Lose (\%) & Tie (\%) & & Win (\%) & Lose (\%) & Tie (\%) & \\
\hline
ARAML vs. MLE & 56.5** & 36.5 & 7.0 & 0.456 & 50.5** & 37.5 & 12.0 & 0.465 \\
ARAML vs. RAML & 54.5** & 37.5 & 8.0 & 0.416 & 47.0* & 40.5 & 12.5 & 0.480  \\
ARAML vs. DialogGAN & 75.5** & 13.5 & 11.0 & 0.445 & 73.0** & 11.0 & 16.0 & 0.460  \\
ARAML vs. DPGAN & 57.5** & 36.0 & 6.5 & 0.435 & 56.5** & 30.5 & 13.0 & 0.529 \\
\hline
\end{tabular}}
\caption{Human evaluation on WeiboDial.
The scores represent the percentages of \textit{Win}, \textit{Lose} or \textit{Tie} when our model is compared with a baseline. $\kappa$ denotes Fleiss' kappa (all are {\it moderate agreement}). The scores marked with * mean \textit{p-value}$<0.05$ and ** indicates \textit{p-value}$<0.01$ in sign test.}
\label{tab:weibohuman}
\end{table*}

\subsection{Dialogue Generation on WeiboDial}



Dialogue evaluation is an open problem and existing works have found that automatic metrics have low correlation to human evaluation \cite{Liu2016Hownot,Novikova17whyweneed,chaganty2018price}. Thus, we resorted to manual evaluation to assess the generation quality on WeiboDial. We randomly sampled 200 posts from the test set and collected the generated results from all the models. For each pair of responses (one from ARAML and the other from a baseline, given the same input post), five annotators were hired to label which response is better (i.e. win, lose or tie) in terms of \textit{grammaticality} (whether a response itself is grammatical and logical) and \textit{relevance} (whether a response is appropriate and relevant to the post). The two metrics were evaluated independently.

The evaluation results are shown in Table \ref{tab:weibohuman}. To measure the inter-annotator agreement, we calculated Fleiss' kappa \cite{fleiss1971kappa} for each pair-wise comparison where results show {\it moderate agreement} ($0.4\leq\kappa\leq0.6$). We also conducted sign test to check the significance of the differences.

As shown in Table \ref{tab:weibohuman}, ARAML performs significantly better than other baselines in all the cases. This result indicates that the samples surrounding true responses provide stable rewards for the generator, and stable RAML training paradigm significantly enhances the performance in both metrics.

\subsection{Further Analysis on Stability}

\begin{figure}[!htp]
  \centering
  \includegraphics[width=1.0\linewidth]{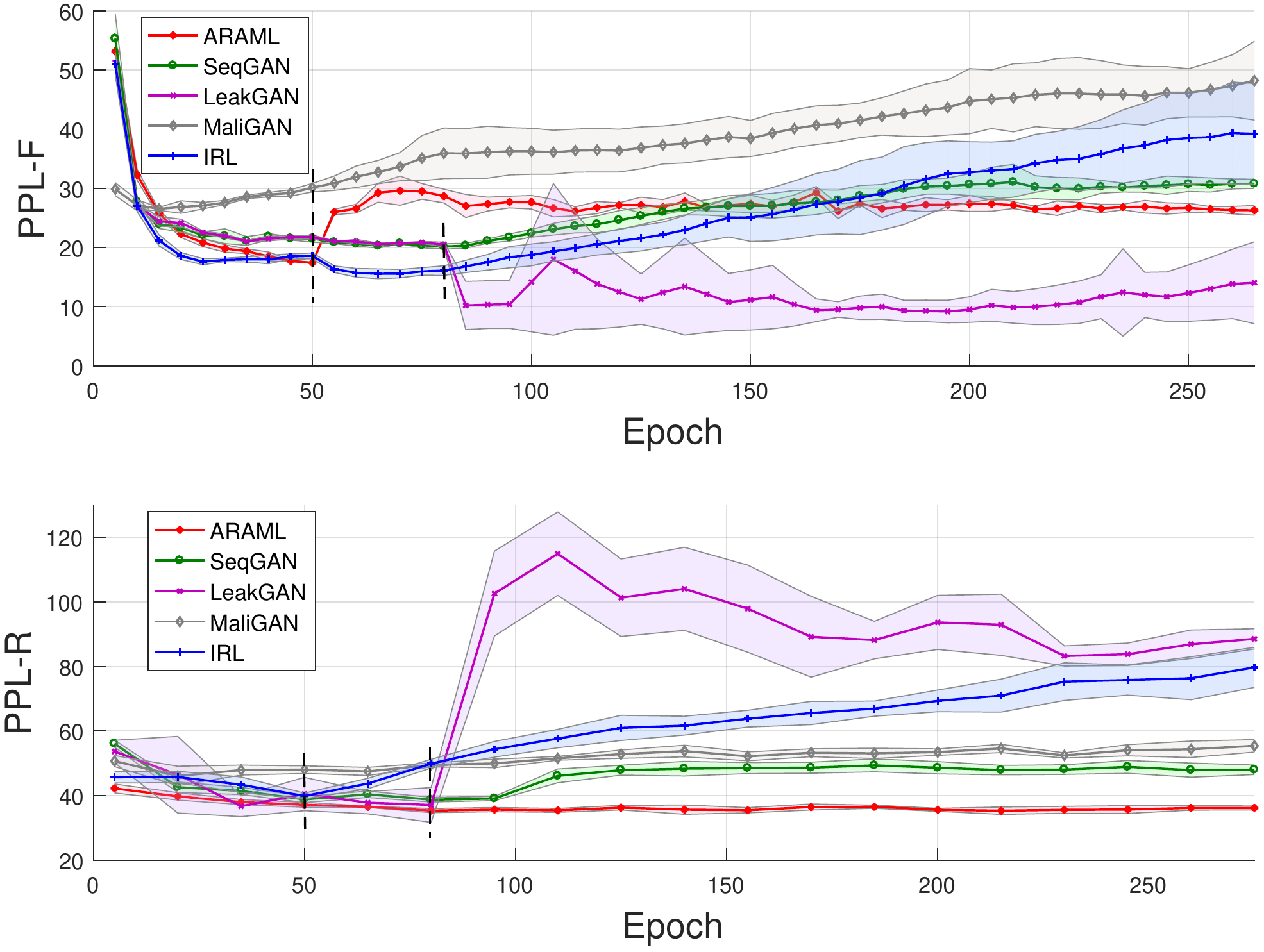}
  \caption{PPL-F/PPL-R curves of ARAML, SeqGAN, LeakGAN, MaliGAN and IRL in the training process. The shade area indicates the standard deviation at each data point. The dotted vertical lines separate pre-training and adversarial training phases (50 for ARAML, IRL and MaliGAN, 80 for SeqGAN and LeakGAN). }
  \label{fig:internal}
\end{figure} 

To verify the training stability, we conducted experiments on COCO many times and chose the best 5 trials for SeqGAN, LeakGAN, IRL, MaliGAN and ARAML, respectively. Then, we presented the forward/reverse perplexity in the training process in Figure \ref{fig:internal}. We can see that our model with smaller standard deviation is more stable than other GAN baselines in both metrics. Although LeakGAN reaches the best forward perplexity, its standard deviation is extremely large and it performs badly in reverse perplexity, indicating that it generates limited expressions that are grammatical yet divergent from the data distribution. 

\subsection{Ablation Study}

\subsubsection{Impact of Temperature}

The temperature $\tau$ controls the search space surrounding the real data as we analyze in Section \ref{sec:editdis}. To investigate its impact on the performance of our model, we fixed all the other hyper-parameters and test ARAML with different temperatures on COCO.

\begin{figure}[!htp]
  \centering
  
  \subfigure{
  \begin{minipage}[t]{1.0\linewidth}
  \centering
  \includegraphics[width=1.0\linewidth]{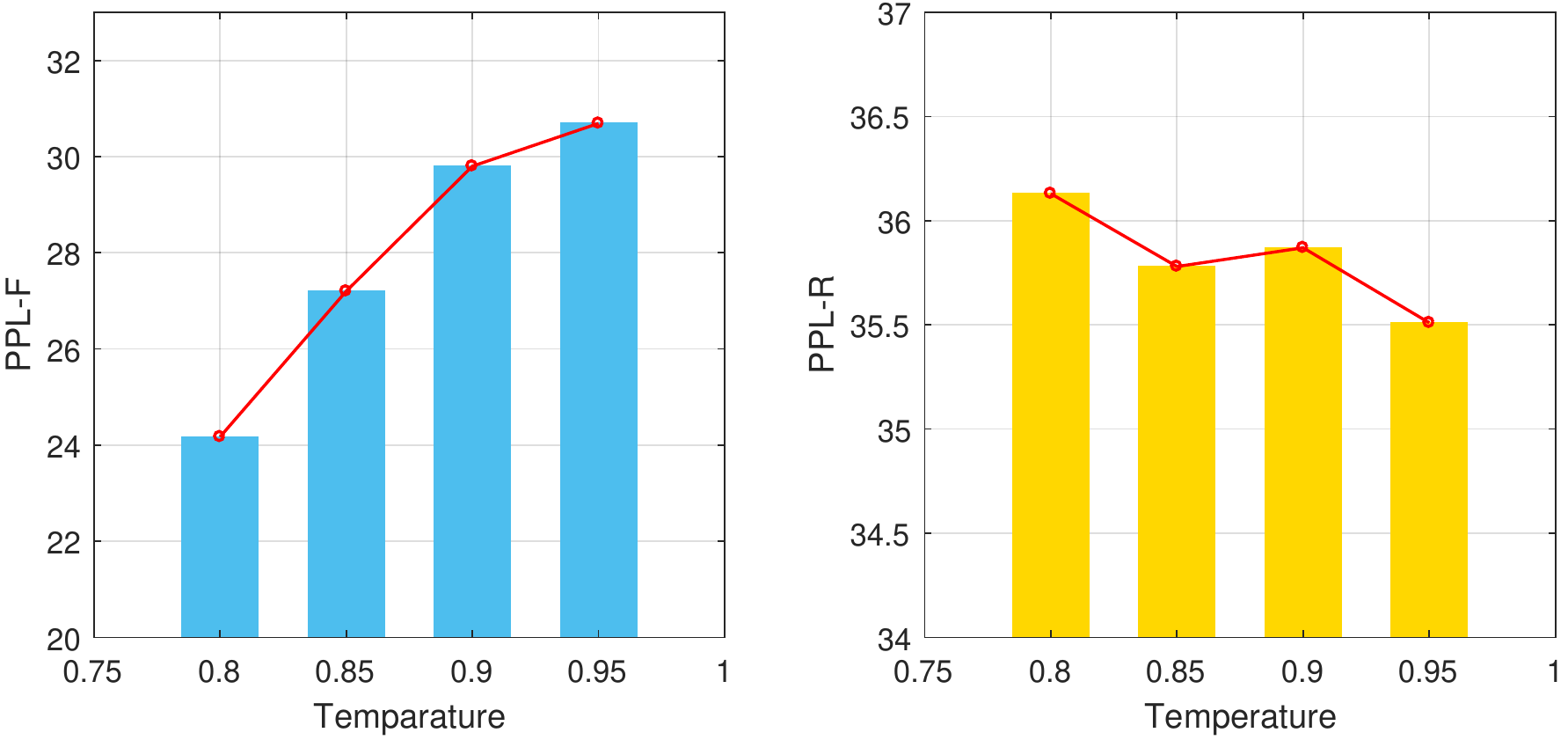}
  \end{minipage}
  }
  
  \subfigure{
  \begin{minipage}[t]{1.0\linewidth}
  \centering
  \includegraphics[width=1.0\linewidth]{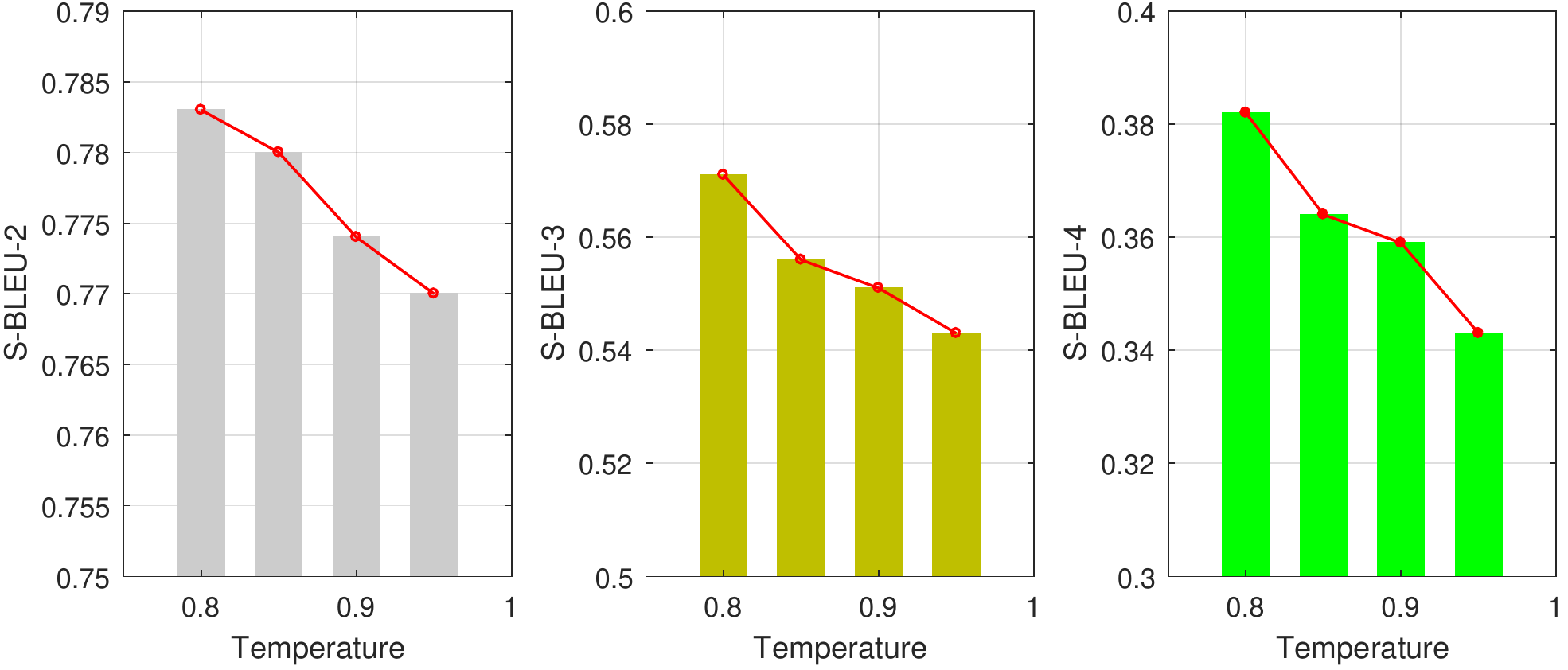}
  \end{minipage}
  }
  
  \centering
  \caption{PPL-F, PPL-R and S-BLEU of ARAML with different temperatures $\tau \in \{0.8,0.85,0.9,0.95\}$ on COCO.}
  \label{fig:temperature}
\end{figure}

The experimental results are shown in Figure \ref{fig:temperature}. We can see that as the temperature becomes larger, forward perplexity increases gradually while Self-BLEU decreases. As mentioned in Section \ref{sec:editdis}, large temperatures encourage our generator to explore the samples that are distant from real data distribution, thus the diversity of generated results will be improved. However, these samples distant from the data distribution are more likely to be poor in fluency, leading to worse forward perplexity. Reverse perplexity is influenced by both generation quality and diversity, so the correlation between temperature and reverse perplexity is not intuitive. We can observe that the model with $\tau=0.95$ reaches the best reverse perplexity.

\subsubsection{Impact of Sampling Strategy}

We have mentioned two common sampling strategies in Section \ref{sec:editdis}, i.e. random sampling and constrained sampling. To analyze their impact, we keep all the model structures and hyper-parameters fixed and test ARAML with these two strategies on COCO.

\begin{table} [!htp]
\centering
\small
\setlength{\tabcolsep}{1.0mm}{
\begin{tabular}{cccc}
\hline
Model  & PPL-F & PPL-R & S-BLEU-2/3/4 \\
\hline
ARAML-R & 37.48$\pm$0.53 & 37.44$\pm$0.56 & \textbf{0.752}/0.571/0.384\\
ARAML-C & \textbf{26.97$\pm$0.55} & \textbf{35.79$\pm$0.49} & 0.777/\textbf{0.560}/\textbf{0.366} \\
\hline
\end{tabular}}
\caption{PPL-F, PPL-R and S-BLEU of ARAML with random sampling (ARAML-R) and constrained sampling (ARAML-C) on COCO.}
\label{tab:rancons}
\end{table}

Table \ref{tab:rancons} shows the results. It's obvious that random sampling hurts the model performance except Self-BLEU-1, because it indeed allows low-quality samples available to the generator. Exploring these samples degrades the quality and diversity of generated results. Despite the worse performance on automatic metrics, random sampling doesn't affect the training stability of our framework. The standard deviation of ARAML-R is still smaller than other GAN baselines.

\subsection{Case Study}

\begin{table} [!htp]
\centering
\small
\setlength{\tabcolsep}{0.8mm}{
\begin{tabu}{l|l}
\hline
Model & Generated Samples \\
\hline
MLE & A little girl sitting on a beach \textcolor{red}{in front of flying} \\
& \textcolor{red}{her kite} at the beach. \\
& A little boy standing in a room next to a desk. \\
\hline
SeqGAN & A man sitting on a bench with snow board \\
& in the background. \\
 & A brown gray cat is in the corner of a street.\\
\hline
LeakGAN & A person that is holding something while \\
& another kid is standing in the water. \\
& A room with a television, mantle, and a chair. \\
\hline
MaliGAN & A man with a shirt on holding one large pink \\
& giant and white kite. \\
& A couple and vases are outside on the bed.\\
\hline
IRL &  \textcolor{blue}{A group of people} wearing helmet \textcolor{green}{sitting down} \\
& \textcolor{green}{on a cell phone}.\\
& \textcolor{blue}{A group of people} sitting in the middle of tracks. \\
\hline
\hline
ARAML & A man is wearing a hat and holding a toothbrush \\
 & as he stands on the grass of a field. \\
& A boy reading a book on a sofa in a room. \\
\hline

\end{tabu}}
\caption{Examples of generated sentences on COCO. Grammatical errors are in red, while blue text represents repetitive expressions and green part indicates incoherent statements.}
\label{tab:cococase}
\end{table}



Table \ref{tab:cococase} presents the examples generated by the models on COCO. We can find that other baselines suffer from grammatical errors (e.g. {\it ``in front of flying her kite"} from MLE), repetitive expressions (e.g. {\it ``A group of people"} from IRL) and incoherent statements (e.g. {\it ``A group of people sitting on a cell phone''} from IRL). By contrast, our model performs well in these sentences and has the ability to generate grammatical and coherent results.

\begin{table} [!htp]
\centering
\footnotesize
\setlength{\tabcolsep}{0.8mm}{
\begin{tabu}{ll}
\hline
Post: & 你还迟到哪？你没这习惯啊 \\
& Are you late? You don’t have this habit. \\
\hline
MLE: & 你是不是要好好照顾自己的！！！ \\
& Would you take care of yourself !!! \\
RAML: &  这是我的生活，我也不想干 \\
& This is my life. I don’t want to do this. \\
DialogGAN: & 这社会的温度快到了。周末，要躺床上！ \\
& The temperature of the society has nearly  \\
& reached. I’ll lie on my bed on the weekend! \\
DPGAN: & 我在上班，我就不习惯了 \\
& I’m working and I’m used to doing this. \\
\hline
\hline
ARAML:  & 嗯，周末就不迟到了！ \\
& Hum. I won’t be late on the weekend! \\
\hline

\end{tabu}}
\caption{Examples of generated responses on WeiboDial.}
\label{tab:weibocase}
\end{table}

Table \ref{tab:weibocase} shows the generated examples on WeiboDial. It's obvious that other baselines don't capture the topic word ``late" in the post, thus generate irrelevant responses. ARAML can provide a response that is grammatical and closely relevant to the post.

\section{Conclusion}

We propose a novel adversarial training framework to deal with the instability problem of current GANs for text generation. 
To address the instability issue caused by policy gradient, we incorporate RAML into the advesarial training paradigm to make our generator acquire stable rewards. Experiments show that our model performs better than several state-of-the-art GAN baselines with lower training variance, yet producing better performance on three text generation tasks.

\section*{Acknowledgments}
This work was supported by the National Science Foundation of China (Grant No. 61936010/61876096) and the National Key R\&D Program of China (Grant No. 2018YFC0830200). We would like to thank THUNUS NExT Joint-Lab for the support.

\bibliography{emnlp-ijcnlp-2019}
\bibliographystyle{acl_natbib}

\end{CJK*}
\end{document}